\documentclass[conference]{IEEEtran}
\IEEEoverridecommandlockouts
\usepackage[ruled,linesnumbered]{algorithm2e}
\usepackage{cite}
\usepackage{amsmath,amssymb,amsfonts}
\usepackage{algorithmic}
\usepackage{graphicx}
\usepackage{textcomp}
\usepackage{xcolor}

\usepackage{multirow}
\usepackage{enumitem}
\usepackage{amsthm}
\usepackage{subfigure}

\theoremstyle{definition}
\newtheorem{definition}{Definition}[]

\def\BibTeX{{\rm B\kern-.05em{\sc i\kern-.025em b}\kern-.08em
    T\kern-.1667em\lower.7ex\hbox{E}\kern-.125emX}}
\begin{document}

\title{AdapTraj: A Multi-Source Domain Generalization Framework for Multi-Agent Trajectory Prediction\\
}

\author{
\IEEEauthorblockN{
Tangwen Qian\textsuperscript{\rm 1,3}, 
Yile Chen\textsuperscript{\rm 2}\IEEEauthorrefmark{1}, 
Gao Cong\textsuperscript{\rm 2}, 
Yongjun Xu\textsuperscript{\rm 1,3}, 
Fei Wang\textsuperscript{\rm 1,3}\IEEEauthorrefmark{1}}
\IEEEauthorblockA{
\textsuperscript{\rm 1}\textit{Institute of Computing Technology, Chinese Academy of Sciences, Beijing, China} \\
\textsuperscript{\rm 2}\textit{School of Computer Science and Engineering, Nanyang Technological University, Singapore}\\
\textsuperscript{\rm 3}\textit{University of Chinese Academy of Sciences, Beijing, China}\\
\{qiantangwen, xyj, wangfei\}@ict.ac.cn, 
yile001@e.ntu.edu.sg, 
gaocong@ntu.edu.sg
}
\thanks{\IEEEauthorrefmark{1} Corresponding Author.}
}

\maketitle

\begin{abstract}
    Multi-agent trajectory prediction, as a critical task in modeling complex interactions of objects in dynamic systems, has attracted significant research attention in recent years. Despite the promising advances, existing studies all follow the assumption that data distribution observed during model learning matches that encountered in real-world deployments. However, this assumption often does not hold in practice, as inherent distribution shifts might exist in the mobility patterns for deployment environments, thus leading to poor domain generalization and performance degradation. Consequently, it is appealing to leverage trajectories from multiple source domains to mitigate such discrepancies for multi-agent trajectory prediction task. However, the development of multi-source domain generalization in this task presents two notable issues: (1) negative transfer; (2) inadequate modeling for external factors. To address these issues, we propose a new causal formulation to explicitly model four types of features: domain-invariant and domain-specific features for both the focal agent and neighboring agents. Building upon the new formulation, we propose AdapTraj, a multi-source domain generalization framework specifically tailored for multi-agent trajectory prediction. AdapTraj serves as a plug-and-play module that is adaptable to a variety of models. 
    Extensive experiments on four datasets with different domains demonstrate that AdapTraj consistently outperforms other baselines by a substantial margin.
\end{abstract}


\section{Introduction}\label{sec:intro}
    As a critical and fundamental task for planning and tracking in autonomous systems under dynamic environments, trajectory prediction of moving objects (e.g., cars, pedestrians, and cyclists) has attracted significant research attention. 
    Numerous methods have been proposed to tackle this problem with a notable emphasis on human motion forecasting due to its prevalence in various downstream scenarios~\cite{Cong-Survey}. Apart from the consideration of individual mobility patterns derived from past trajectories, the interactions among objects, such as collision avoidance and gathering behaviors, also exert substantial influence on human motion. To this end, recent studies~\cite{SocialLSTM,SocialGAN,STGAT,PECNet} have demonstrated the effectiveness of incorporating the interaction modeling among different objects (i.e., agents) in this task, which is commonly referred to as multi-agent trajectory prediction. 

    Despite the prominent results achieved by these studies, they all assume the same data distribution between the training and testing stages. In other words, the trained models can be seamlessly deployed in environments that have similar characteristics to the training instances. However, this assumption often does not hold in practice, as inherent distribution shifts might exist in the mobility patterns for dynamic ever-changing deployment environments. To quantitatively illustrate the presence of such distribution shifts, we present the statistics for four commonly utilized human trajectory datasets in Tab.~\ref{tab: details_of_datasets}. We observe significant discrepancies in human motion characteristics (e.g., crowd density (num), velocity (v), and acceleration(a)) across them. For instance, the SYI dataset exhibits the highest average velocity and acceleration on y-axis, which is approximately 26 and 7 times greater than that of the L-CAS dataset respectively. 

    \begin{table}[tbp]
    \centering
        \caption{
            Statistical analysis of four distinct datasets.
            We present the average and standard deviation for several trajectory characteristics along horizontal (x) and vertical (y) axes.
        }
        \vspace{-3mm}
        \resizebox{0.96\linewidth}{!}{
            \begin{tabular}{c|c|c|c|c}
                \hline
                Datasets        & ETH\&UCY & L-CAS & SYI & SDD \\
                \hline \hline
                \# sequences            & 3856         & 2499        & 5152          & 35634 \\
        Avg/Std num          & 9.09/10.01 & 7.88/3.23 & 35.17/20.81 & 17.82/15.12 \\
        Avg/Std v(x)    & 0.279/0.170  & 0.104/0.078 & 0.306/0.063   & 0.295/0.204\\
        Avg/Std v(y)    & 0.090/0.070  & 0.041/0.024 & 1.087/0.185   & 0.187/0.156 \\
        Avg/Std a(x) & 0.027/0.027  & 0.044/0.028 & 0.082/0.018   & 0.057/0.042 \\
        Avg/Std a(y) & 0.027/0.024  & 0.044/0.025 & 0.339/0.062   & 0.064/0.053 \\
                \hline
            \end{tabular}}
            \label{tab: details_of_datasets}
        \vspace{-9mm}
    \end{table}

    As a result, such a distribution shift issue poses a significant challenge to existing methods for multi-agent trajectory prediction. 
    Existing methods exhibit a significant decline in performance when applied to the instances from a different domain (details are in Sec.~\ref{subsec: motivation}). 
    It seems that training a distinct model for each domain could serve as a solution to this issue. However, this approach is neither efficient nor feasible given the diversity of scenarios and mobility patterns in real use cases. 
    In particular, given that not every environment will be equipped with sensors or cameras to record trajectories, it is unlikely to have trained models for every scenario or domain. Moreover, even when sufficient training data is available, this idea becomes progressively intractable. The increased number of domains would lead to the unbearable growth in the model sizes and the cost of training time.
    Therefore, this indicates the necessity for a generic model that can handle trajectories with varying characteristics, and even those unseen domains that have not been previously encountered in training instances. 

    In light of this, limited research has been done to address the challenge posed by the distribution shift issue. For example, Counter~\cite{Counter} and CausalMotion~\cite{CausalMotion} are devised to learn a transferable model from trajectories in one domain and retain its effectiveness when applied to another domain, thereby achieving single-source domain generalization.
    Unfortunately, 
    the relatively limited number of instances available for a single domain, coupled with the substantial discrepancies in trajectory characteristics between the source and the target domain, makes it difficult for these methods to sufficiently encode the required knowledge about the variations present in target domains. 
    Therefore, it is appealing to leverage trajectories from multiple source domains for multi-agent trajectory prediction problem for two reasons: (1) since deep learning models are usually data-hungry for capturing the domain knowledge, a larger number of trajectories provide more opportunities to model a variety of influential factors (e.g., environment bias, and diverse motion patterns) required for effective performance. Moreover, the presence of trajectories characterized by different distributions offers the opportunity to better capture both the invariant features and the specific features; 
    (2) despite the discrepancies in trajectory characteristics across domains, most domains are correlated with each other~\cite{Meta-DMoE,generalizationSurvey,mixture-of-experts} in the sense that they all adhere to certain intrinsic patterns which are implicitly hidden within mobility regardless of the domain. These patterns include the immutable physical laws and the conventions of social norms, and influence how people move and interact in space~\cite{SocialForce, SocialLSTM,SocialGAN}. 
    However, it is still challenging to apply multi-source domain generalization due to two issues.

    The first issue is the occurrence of \emph{negative transfer} from multiple source domains.  While multiple source domains can provide an increased quantity of data instances, it is usually problematic to directly fuse the training data from these sources to learn invariant mobility patterns across domains. Instead, without any constraints, the model tends to simply average the discrepancies in multiple source domains, which usually leads to inferior model performance (details are in Sec.~\ref{subsec: motivation}).

    The second issue is the \emph{inadequate modeling for external factors}. Counter~\cite{Counter} observes an obvious gap of external factors among different domains and proposes a counterfactual analysis method to remove the dependence of external factors (e.g., influences from neighboring agents) and only utilize the invariant features in individual trajectories. Nevertheless, human motion is not only decided by internal initiatives, but also greatly affected by external factors, especially the movements of surrounding agents. In this case, while noise features are eliminated, this method also ignores some reasonable influences hidden in external factors. Moreover, CausalMotion~\cite{CausalMotion} endeavors to model external factors from a causal perspective. However, it is designed to leverage only a single source domain, thus not capable of uncovering the invariant features shared by external factors that are common across multiple domains. 

    To address these two issues, we propose a multi-source domain generalization framework tailored for multi-agent trajectory prediction, named AdapTraj, which serves as a plug-and-play module that is \emph{adaptable} to a variety of models. 
    To mitigate the first issue,
    apart from obtaining invariant features that are consistent across domains, we aim to \emph{adaptively} aggregate the domain-specific features based on the discrepancies within each source domain, thus capturing a more diverse range of personalized patterns overlooked in invariant features. In particular, we employ extractors with orthogonal constraints to split the knowledge contained in each domain into domain-invariant and domain-specific features. Leveraging these disentangled features, we devise a teacher-student learning process where the proposed aggregator can utilize collective experiences from the trained extractors to obtain effective domain-specific features to accommodate trajectories presented by unseen domains during inference. Furthermore, 
    to address the second issue,
    as opposed to the oversimplified assumption employed in~\cite{Counter, CausalMotion}, we propose a new formulation that considers both the domain-invariant features (e.g., collision avoidance and gathering behaviors) and domain-specific features (e.g., yielding right-of-way or left-of-way) of neighboring agents, which are the dominating external factors. We devise extractors to produce these two types of features, which are then fused with the corresponding features derived from the focal agent to serve as the distilled knowledge from multiple source domains. Finally, given a new trajectory from an unseen domain, the future trajectory is generated by integrating its historical information and neighbor interactions, along with the two types of features that encode the knowledge about the invariant and diverse patterns across domains. Our model serves as a holistic solution which integrates principles of domain generalization while addressing the distinct requirements of multi-agent trajectory prediction task. 

    The contributions are summarized as follows:
    \begin{itemize}[leftmargin=*]
        \item We identify the limitations through quantitative experiments 
        (i.e., Tab.~\ref{tab: domain generalization performance drop} and Tab.~\ref{tab: negative transfer})
        on how to obtain a generalizable model that is effective at tackling unseen examples in multi-agent trajectory prediction task. Consequently, we propose a plug-and-play framework, named AdapTraj, that can be incorporated into existing
        deep learning based methods to harness the knowledge from multiple domains, thus achieving better generalization performance in unseen domains. 
        \item The proposed AdapTraj is
        designed to effectively derive
        the domain-invariant features and the domain-specific features for both the focal and the neighboring agents. These features encode not only the knowledge shared across domains, but also the diverse patterns required to handle the variations present in different domains. Based on these features, we further devise a new causal formalism to predict future trajectories.
        \item We implement the proposed method on two widely used state-of-the-art methods for multi-agent trajectory prediction task. 
        Extensive experiments on four real-world datasets from different domains 
        show that our proposed method consistently outperforms  baselines by a substantial margin.
    \end{itemize}

\vspace{-2mm}
\section{Preliminaries}

    In this section, we present the problem formulation for our problem, 
     detailed observations for our motivation,
    and introduce some preliminaries to understand the concepts for conventional multi-agent trajectory prediction task.

\subsection{Problem Definition}
\label{subsec: problem definition}

    We start with introducing some definitions employed in the context of multi-agent trajectory prediction and multi-source domain generalization. Subsequently, we provide a formal definition of the research problem studied in this paper. 

    \begin{definition}(\textbf{Trajectory}).
        A trajectory of length $\vert T \vert$ is defined as a sequence of  locations $X_i^{T} = \{ (lon_i^t, lat_i^t) \}_{t=1}^{T}$ travelled by an agent, where $lon_i^t$ and $lat_i^t$ denote longitude and latitude of the $i$-th agent at timestamp $t$.
    \end{definition}

    \begin{definition}(\textbf{Multi-Agent Trajectory Prediction}).
        Given the observed trajectory of a focal agent $X_i^{\tau} = \{ (lon_i^t, lat_i^t) \}_{t=\tau+1-\vert T_{obs} \vert}^{\tau}$ in the past $|T_{obs}|$ steps, and the trajectories of its neighboring agents $N(i,\tau)$ 
        that co-occur with the focal agent within a specified time interval
        $\mathcal{E}_{i}=\{X_{j}^{\tau}\}_{j=1}^{N(i,\tau)}$, the objective of multi-agent trajectory prediction can be modeled as learning a predictive function:

        \begin{equation}
           \hat{Y}_i^{\tau} = \mathcal{F}(X_{i}^{\tau}, \mathcal{E}_{i})
           \nonumber
        \end{equation}
        where $\hat{Y}_i^{\tau}=\{ (\hat{lon}_i^t, \hat{lat}_i^t) \}_{t=\tau +1}^{\tau + \vert T_{pred} \vert}$ is the predicted future trajectory of the focal agent in the next $|T_{pred}|$ steps. 
    \end{definition}

    Compared to single-agent prediction, the phrase ``multi-agent trajectory prediction'' highlights the analytical process that accounts for the dynamic interactions among multiple agents. Such interactions lead to diverse movement behaviors, including but not limited to collision avoidance, leader-follower relationships, and coordination based on social norms.

    While additional data such as visual (e.g., environmental images) or contextual information (e.g., group annotations) are sometimes available, in line with studies~\cite{LBEBM,Evolvegraph,SocialLSTM,PECNet} for broader practical applications, trajectories are used as the only input for prediction in our study.

    \begin{definition}(\textbf{Domain Generalization}).
        The objective of domain generalization is to learn a robust predictive function $f: \mathcal{X} \rightarrow \mathcal{Y}$ using solely the data from a collection of source domains $\mathcal{D}_S =\{ D_{S}^1, D_{S}^2, \dots, D_{S}^K \}$ such that the prediction error on unseen target (test) domain $D_{T}$ is minimized: 
        \begin{equation}
            \min\limits_{f} \mathbb{E}_{(\mathbf{x}, \mathbf{y}) \in D_T} [\mathcal{L}(f(\mathbf{x}), \mathbf{y})]
        \nonumber
        \end{equation}
        where $\mathbb{E}$ denotes the expectation and $\mathcal{L}(\cdot, \cdot)$ represents the loss function calculated by the predicted result $f(\mathbf{x})$ and the ground truth $\mathbf{y}$. 
        
        We follow the commonly adopted setting~\cite{generalizationSurvey,generalizationApplication,generalizationSurvey2,DSN} by assuming that the target domain can not be accessed during model training, and the joint distribution $P(X, Y)$ for each domain differs from one another:
        \begin{equation}
            P_T (X, Y) \neq P_{S}^{k} (X, Y), \forall k \in \{ 1,2,\dots, K \}
        \nonumber
        \end{equation}
        \begin{equation}
            P_{S}^{k} (X, Y) \neq P_{S}^{k^{'}} (X, Y), 1 \leq k \neq k^{'} \leq K
        \nonumber
        \end{equation}
    \end{definition}
    
    Note that for multi-agent trajectory prediction task, previous studies~\cite{Counter, CausalMotion} propose to learn $f$ by leveraging training data from a single source domain (i.e., $K=1$), which is referred to as single-source domain generalization. In contrast, we aim to expand the number of source domains as multi-source domain generalization setting, and propose to study how to enhance the performance by effectively utilizing multiple distinct yet correlated source domains (i.e., $K > 1$). 

    Building upon the above definitions, we provide a formal definition of the research problem addressed in this paper. 

    \begin{definition}(\textbf{Multi-Source Domain Generalization for Multi-Agent Trajectory Prediction}). Given trajectory data from multiple domains $\mathcal{D}_{S}$, we aim to learn a predictive function $\mathcal{F}$ that
    minimizes the error of multi-agent trajectory prediction for trajectories from unseen target domain $D_{T}$:
        \begin{equation}
            \min\limits_{\mathcal{F}} \mathbb{E}_{([X_i^t, \mathcal{E}_{i}], Y_i^t) \in D_T} [\mathcal{L}(\mathcal{F}(X_i^t, \mathcal{E}_{i}), Y_i^t)]
        \nonumber
        \end{equation}
    \end{definition}

\subsection{Motivation}
\label{subsec: motivation}

    We illustrate the issues for domain generalization and negative transfer from multiple domains discussed in Sec.~\ref{sec:intro}. 

\subsubsection{Impact of domain discrepancies on performance decline}

    We verify the issue of domain discrepancies with two state-of-the-art methods, namely LBEBM~\cite{LBEBM} and PECNet~\cite{PECNet}. As shown in Tab.~\ref{tab: domain generalization performance drop}, 
    these methods exhibit a significant decline in performance 
    when they are applied to the test instances on the SDD dataset, but are trained on instances from a different domain (ETH\&UCY), in comparison to instances from the same domain (SDD). This motivates us to design effective methods that can tackle various domains for multi-agent trajectory prediction.  

    \begin{table}[tbp]
    \centering
        \caption{
            results of existing methods with different target domain on ADE and FDE metrics. 
        }
        \vspace{-2.5mm}
        \resizebox{0.9\linewidth}{!}{
            \begin{tabular}{c|c|c|c|c}
                \hline
                Source Domain & LBEBM     & PECNet    & Counter   & CausalMotion \\
                \hline
                SDD           & 0.55/0.98 & 0.59/1.05 & 1.34/2.93 & 1.35/2.89 \\
                ETH\&UCY      & 0.85/1.80 & 1.20/1.88 & 1.48/3.03 & 1.56/3.28 \\
                \hline
            \end{tabular}}
        \label{tab: domain generalization performance drop}
        \vspace{-4mm}
    \end{table}

\subsubsection{Impact of negative transfer}

    To quantify the impact of negative transfer, we evaluate the performance of two single-source domain generalization methods~\cite{Counter, CausalMotion} for multi-agent trajectory prediction. As indicated in Tab.~\ref{tab: negative transfer}, rather than benefiting from an increased volume of training data, as the number of source domains increases, we observe a decline in both ADE/FDE metrics (lower is better, definitions are in Sec.~\ref{subsec:exp_setup}) on the unseen domain (SDD). This illustrates the difficulty of multi-source domain generalization posed by the phenomenon of negative transfer.  
    
    \begin{table}[tbp]
    \centering
        \caption{
            results of single-source domain generalization methods trained on varied number of source domains. 
        }
        \vspace{-2.5mm}
        \resizebox{0.75\linewidth}{!}{
            \begin{tabular}{c|c|c}
                \hline
                Source Domains      & Counter   & CausalMotion \\
                \hline
                ETH\&UCY           & 1.48/3.03 & 1.56/3.28 \\
                ETH\&UCY, L-CAS     & 1.57/3.17 & 1.85/3.50 \\
                ETH\&UCY, L-CAS, SYI & 1.77/3.68 & 1.89/3.68 \\
                \hline
            \end{tabular}}
            \label{tab: negative transfer}
        \vspace{-5mm}
    \end{table}

\subsection{Backbone Model for Multi-Agent Trajectory Prediction}
\label{sec: conventional multi-agent tp models}

    \begin{figure*}[tbp]
        \centerline{\includegraphics[width=0.7\linewidth]{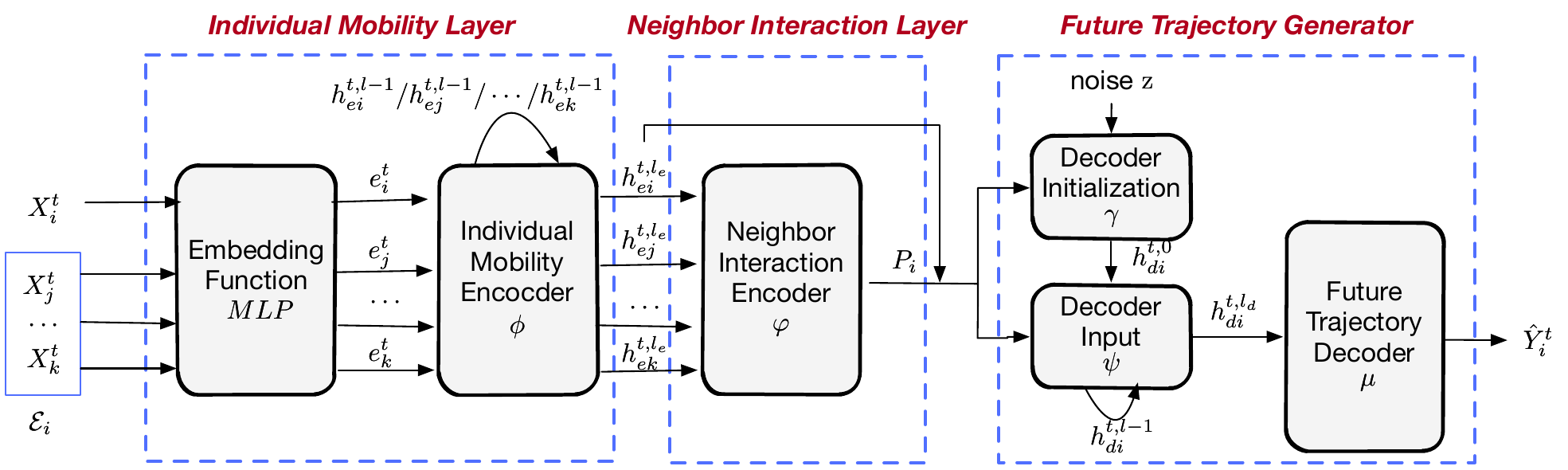}}
        \vspace{-3mm}
        \caption{
        An overview of backbone model.
        }
        \label{fig: base_model_architecture}
    \vspace{-5mm}
    \end{figure*}

    As our framework serves as a plug-and-play module applicable to various multi-agent trajectory prediction methods, we offer a brief overview of the backbone model, which is built upon the sequence-to-sequence (seq2seq) architecture, for conventional multi-agent trajectory prediction task. As illustrated in Fig.~\ref{fig: base_model_architecture}, the model consists of three components: an individual mobility layer, a neighbor interaction layer, and a future trajectory generator. 

\subsubsection{Individual Mobility Layer}
    This layer aims to capture the individual mobility patterns of each agent by utilizing the historical trajectories of the focal agent $X_{i}^{t}$ and its neighboring agents $\mathcal{E}_{i}$ as input. Initially, a multi-layer perception (MLP) is employed to encode the location of each agent, resulting in a fixed-length vector $e^t_i$. The function $MLP(\cdot)$ represents an embedding function with ReLU non-linearity. This encoding process can be expressed as follows: 
    \begin{equation}
        e^t_i = MLP(X^t_i) 
    \end{equation}

    Subsequently, these embeddings serve as input to the individual mobility encoder $\phi(\cdot)$, which captures the state of each agent and extracts its motion pattern. At each iteration $l$ for each agent, denoted as $h^{t, l}_{ei}$, the individual mobility pattern is acquired through the encoder. This encoder, with the number of iterations denoted as $l_e$, can be implemented using any sequential models, such as LSTM~\cite{SocialLSTM,PredRNN}, or more advanced models like Transformer~\cite{TransformerTF,ICDE21-VehiclePredict}. The computation of the encoder can be formulated as: 
    \begin{equation}
        h^{t, l}_{ei} = \phi(h^{t, l-1}_{ei}, e^t_i), 1 \leq l \leq l_{e}
    \end{equation}
    
\subsubsection{Neighbor Interaction Layer}
    The neighbor interaction layer combines the mobility pattern of different agents to effectively model the interactions between the focal agent and its neighboring agents. We aggregate the hidden states of all agents present in the scene to generate an interaction tensor $P_i$ for each agent. The calculation is defined as follows: 
    \begin{equation}
        P_i = \varphi(h^{t, l_e}_{ei}, h^{t, l_e}_{ej}, \cdots, h^{t, l_e}_{ek}), \forall j, k \in N(i,t) 
    \end{equation}
    where $\varphi(\cdot)$ denotes a specifically designed encoder, such as pooling~\cite{SocialLSTM,SocialGAN}, attention\cite{STGAT,PECNet}, and graph mechanism\cite{Evolvegraph,CausalMotion}. $h^{t, l_e}_{ei}$, $h^{t, l_e}_{ej}$, and $h^{t, l_e}_{ek}$ denote the individual mobility pattern of the $i$-th, $j$-th, and the $k$-th agent, respectively, at the final iteration $l_e$. Moreover, the $j$-th and $k$-th agents are neighbors of the $i$-th agent. 

\subsubsection{Future Trajectory Generator}
    The purpose of the future trajectory generator is to generate plausible future trajectories that align with the observed historical trajectories. To achieve this, we adopt the seq2seq decoder architecture initialized by the final hidden state from the encoder to perform the generation of output trajectories. Specifically, the initialization steps are defined as follows:
    \begin{equation}
        c^t_i = \gamma (P_i, h^{t, l_e}_{ei})
    \end{equation}
    \begin{equation}
        h^{t, 0}_{di} = [c^t_i, z]
    \end{equation}
    where $\gamma(\cdot)$ represents a decoder initialization, such as MLP, that combines the final hidden state of individual mobility and neighbors' interaction state. Here $h^{t, 0}_{di}$ corresponds to the initialized decoder state, and $z$, which refers to the latent variable serving as input noise, is utilized to generate multiple diverse yet socially compliant future trajectories. Once the decoder states are initialized as described above, trajectory prediction is carried out using the following steps: 
    \begin{equation}
        h^{t,l}_{di} = \psi ( P_i, h^{t, l_e}_{ei}, h^{t, l-1}_{di}), 1 \leq l \leq l_d
    \end{equation}
    \begin{equation}
        \hat{Y}^t_i = \mu (h^{t, l_d}_{di})
    \end{equation}
    where $\psi(\cdot)$ represents a designed function that combines different features and undergoes $l_d$ iterations to perform its operation. $\mu(\cdot)$ denotes the designed future trajectory decoder responsible for generating the future trajectory based on the combined feature described above. Here $h^{t, l}_{di}$ corresponds to the decoder state of each agent at iteration $l$. 

    The training objective aims to minimize errors between the predicted future trajectories and the ground truth trajectories: 
    \begin{equation}
        L_{base} = \sum_{Y_i^t \in D_S} \Vert Y_i^t - \hat{Y}^t_i \Vert_2^2
    \label{eq: base loss}
    \end{equation} 
    
\section{AdapTraj Framework} 
\vspace{-3mm}

    \begin{figure*}[htbp]
        \centerline{\includegraphics[width=0.7\linewidth]{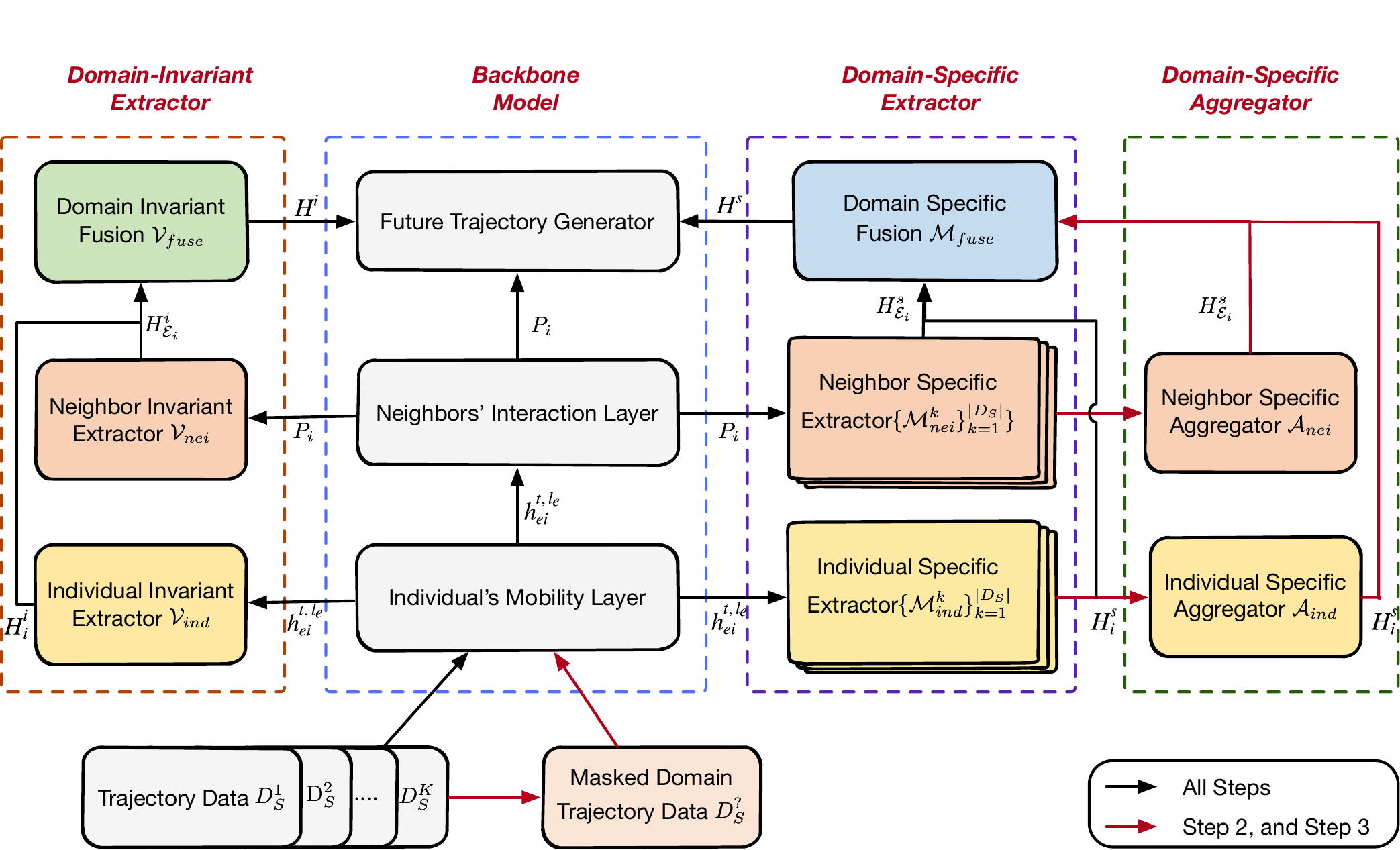}}
        \caption{
        An overview of AdapTraj. AdapTraj consists of three components: domain-invariant extractor, domain-specific extractor, and domain-specific aggregator. 
        }
        \label{fig: architecture}
        \vspace{-6mm}
    \end{figure*}

    \subsection{Overview} 
    The proposed AdapTraj framework, designed for multi-source domain generalization setting in multi-agent trajectory prediction, functions as a plug-and-play module adaptable to a variety of models. This framework diverges from the oversimplified assumptions in prior domain generalization studies~\cite{Counter, CausalMotion}, which cannot fully capture the impact of external factors in this task. Instead, as depicted in Fig.~\ref{fig: architecture}, we propose a new causal formulation for multi-agent trajectory prediction and explicitly model four types of features: domain-invariant and domain-specific features for both the focal agent and neighboring agents. This approach is able to effectively harness the knowledge from multiple domains and thus address the negative transfer issue for prediction in unseen target domains. To this end, AdapTraj comprises three main components: a domain-invariant extractor, a domain-specific extractor, and a domain-specific aggregator.

    To address the issue of inadequate modeling for external factors, based on our causal formalism, domain-invariant extractor and domain-specific extractor are employed to effectively derive the various features mentioned above. Specifically, we ensure the learned features from domain-invariant extractor remain invariant despite domain variations by utilizing weight sharing across various source domains. Furthermore, we ensure the learned features from the domain-specific extractor are specific to each domain by taking advantage of the orthogonal relationship between invariant and specific features.

    To address the issue of negative transfer arising from multiple source domains, the domain-specific aggregator is employed to adaptively aggregate the specific knowledge obtained from all source domains and extract effective domain-specific features to accommodate trajectories presented by unseen domains during inference.

    Next, we introduce the details of each module, namely domain-invariant extractor (Sec.~\ref{sec: invariant extractor}), domain-specific extractor (Sec.~\ref{sec: specific extractor}), and domain-specific aggregator (Sec.~\ref{sec: specific aggregator}). We further illustrate the training and inference procedures (Sec.~\ref{sec: training and inference procedure}). 

\subsection{Domain-Invariant Extractor}
\label{sec: invariant extractor}

    Previous studies, as indicated in~\cite{generalizationSurvey,generalizationSurvey2,Meta-DMoE,DSN}, suggest the importance of extracting inherent and domain-invariant knowledge for effective domain generalization. In this way, features that remain invariant to the discrepancies across source domains are more likely to exhibit robustness to any unseen target domain. Applying this concept to the task of multi-agent trajectory prediction, the extraction of invariant features from both the focal agent's and neighbors' trajectories becomes critical as these features function as essential components in the predictive model. Building upon this idea, and drawing inspiration from classical transfer learning work~\cite{generalizationSurvey,generalizationApplication,spatialMoE,self-supervisedDG}, we propose a shared-weight domain-invariant extractor module to capture shared features across various domains. 
    
    Specifically, the domain-invariant extractor module consists of three parts: an individual invariant extractor, a neighbor invariant extractor, and a domain invariant fusion. The individual invariant extractor $\mathcal{V}_{ind}$ is designed to extract invariant features $H_i^i$ from the individual mobility of the focal agent (e.g., human motion), while the neighbor invariant extractor $\mathcal{V}_{nei}$ focuses on extracting invariant features $H_{\mathcal{E}_{i}}^i$ from the interactions of its neighbors (e.g., collision avoidance). The domain invariant fusion $\mathcal{V}_{fuse}$ combines these two invariant features into a unified invariant variable $H^i$ for the prediction of future trajectories.
    \begin{equation}
        H_i^i = \mathcal{V}_{ind} (h_{ei}^{t, l_e})
    \end{equation}
    \begin{equation}
        H_{\mathcal{E}_{i}}^i = \mathcal{V}_{ind} (P_i)
    \end{equation}
    \begin{equation}
        H^i = \mathcal{V}_{fuse} (H_i^i, H_{\mathcal{E}_{i}}^i)
    \end{equation}

    To ensure effective extraction of invariant features, our approach incorporates two loss functions: the reconstruction loss and the domain adversarial similarity loss. Additionally, we introduce two extra decoders, i.e., the reconstruction decoder $D_{recon}$ and domain classifier $D_{class}$, to compute these losses. 

    The \emph{reconstruction loss} employs the scale-invariant mean squared error\cite{simse,DSN} across all source domains. By penalizing differences between pairs of dimensions, the scale-invariant mean squared error allows the model to learn to reproduce the overall shape of the objects being modeled. The reconstruction loss $L_{recon}$ is calculated as follows: 
    \begin{equation}
        L_{recon} = \sum_{X_i^t \in D_S} L_{simse} (X_i^t, \hat{X}_i^t)
    \label{eq: recon loss}
    \end{equation}
    where $\hat{X}_i^t$, the reconstruction of $X_i^t$, is generated by the reconstruction decoder $D_{recon}$, e.g., MLP, using the invariant $H_i^i$ and specific $H_i^s$ features associated with $X_i^t$. 
    \begin{equation}
        \hat{X}_i^t = D_{recon} (H_i^i, H_i^s)
    \end{equation}
    The scale-invariant mean squared error is a metric that relates to the conventional mean squared error but incorporates an additional term $ - \frac{1}{m^2} \Vert X_i^t - \hat{X}_i^t \Vert_2^2$ to credit mistakes if they are in the same direction and penalizes them if they oppose. The computation is as follows: 
    \begin{equation}
        L_{simse}(X_i^t, \hat{X}_i^t) = \frac{1}{m} \Vert X_i^t - \hat{X}_i^t \Vert_2^2 - \frac{1}{m^2} \Vert X_i^t - \hat{X}_i^t \Vert_2^2
    \end{equation}
    where $m$ represents the number of dimensions in input $X_i^t$; and $\Vert \cdot \Vert_2^2$ denotes the squared $L_2$ norm. 

    The \emph{domain adversarial similarity loss} aims to minimize the negative log-likelihood of the ground truth domain label for each sample from source domains. This encourages the model to generate representations in such a way that a domain classifier $D_{class}$ reliably predicts the domain of the encoded representation. The domain adversarial similarity loss $L_{similar}$ is calculated as follows: 
    \begin{equation}
        L_{similar} = \sum_{X_i^t \in D_S} - d_i log(\hat{d}_i)
    \label{eq: similar loss}
    \end{equation}
    where $d_i$ represents the one-hot encoding of the domain label for source input $X_i^t$ and $\hat{d}_i$ is the softmax prediction of the domain label, generated by the domain classifier $D_{class}$.
    \begin{equation}
        \hat{d}_i = D_{class} ( H_i^i, H_{\mathcal{E}_i}^i, H_i^s, H_{\mathcal{E}_i}^s )
    \end{equation}

\subsection{Domain-Specific Extractor}
\label{sec: specific extractor}

    To address the issue of inadequate modeling for external factors, we propose a new causal formulation compared to the oversimplified assumption in prior domain generalization studies. Apart from domain invariant features, in the context of multi-source domain generalization, the incorporation of domain-specific features significantly contributes to enhancing the performance and adaptability of models across different source domains~\cite{mixture-of-experts,Meta-DMoE,spatialMoE}. While domain-invariant features capture the shared underlying patterns and knowledge among various domains, domain-specific features account for the unique characteristics specific to individual domains. Consequently, the extraction of specific features from both the trajectory of the focal agent and the trajectories of its neighbors becomes of paramount importance.

    There exists an orthogonality constraint between domain-specific and domain-invariant features~\cite{DSN,generalizationSurvey,generalizationSurvey2}. Building upon this constraint and drawing inspiration from the concept of mixture-of-experts~\cite{Meta-DMoE,mixture-of-experts,spatialMoE,ICDE21-SoftMoE}, we propose the domain-specific extractor to encourage a clear separation between the features related to specific source domains and the features shared across domains.

    The domain-specific extractor consists of three parts: an individual specific extractor, a neighbor specific extractor, and a domain specific fusion. The individual specific extractor $\mathcal{M}_{ind}$ is designed to extract specific features $H_i^s$ from the individual's mobility, while the neighbor specific extractor $\mathcal{M}_{nei}$ focuses on extracting specific features $H_{\mathcal{E}_i}^s$ from the neighbors' interaction. The domain specific fusion $\mathcal{M}_{fuse}$ combines these two specific features into a unified specific variable $H^s$ which is used for the prediction of future trajectories. 

    Specifically, we define $\mathcal{M}_{ind} = \{ \mathcal{M}_{ind}^k \}_{k=1}^{\vert D_S \vert}$ and $\mathcal{M}_{nei} = \{ \mathcal{M}_{nei}^k \}_{k=1}^{\vert D_S \vert}$ to represent the individual specific extractor and neighbor specific extractor, respectively. In contrast to the domain-invariant extractor module that shares the same weights across source domains, each $\mathcal{M}_{ind}^k$ or $\mathcal{M}_{nei}^k$ is trained separately on the source domain $D_S^k$ to learn its specific features related to the individual or neighbors as follow: 
    \begin{equation}
         H_i^{s} = \mathcal{M}_{ind}^k ( h_{ei}^{t, l_e} ), 1 \leq k \leq \vert D_S \vert
    \end{equation}
    \begin{equation}
         H_{\mathcal{E}_i}^{s} = \mathcal{M}_{nei}^k ( P_i ), 1 \leq k \leq \vert D_S \vert
    \end{equation}
    \begin{equation}
        H^s = \mathcal{M}_{fuse} (H_i^s, H_{\mathcal{E}_{i}}^s)
    \end{equation}

    To ensure that only domain-specific knowledge is encoded, our method integrates the \emph{difference loss} $L_{diff}$ via a soft subspace orthogonality constraint between specific $H_i^s$ and invariant $H_i^i$ representations of each domain. 
    \begin{equation}
        L_{diff} = \sum_{X_i^t \in D_S} \Vert {H_i^i}^\mathrm{T} H_i^s \Vert_F^2 +  \Vert {H_{\mathcal{E}_i}^i}^\mathrm{T} H_{\mathcal{E}_i}^s \Vert_F^2
    \label{eq: diff loss}
    \end{equation}
    where $\Vert \cdot \Vert_F^2$ denotes the squared Frobenius norm. 

    In addition to the difference loss $L_{diff}$, we employ $L_{recon}$ and $L_{similar}$ as mentioned in the domain-invariant extractor module to enhance effectiveness of extracted specific features. 

\subsection{Domain-Specific Aggregator}
\label{sec: specific aggregator}

    To tackle the issue of negative transfer arising from multiple source domains and to enhance the model's adaptability in the unseen target domain, the domain-specific aggregator is designed. In the context of multi-source domain generalization setting, where the target domain cannot be accessed during both training and evaluation stages, we simulate the test-time distribution shift by excluding the corresponding expert model in each iteration. This simulation is achieved by masking out the domain label ($D_S^k \rightarrow D_S^?$ in Fig.~\ref{fig: architecture}). Consequently, the domain-specific aggregator is compelled to adapt using the aggregated domain-specific knowledge obtained from the accessible source domains, leading to improved generalization performance in the unseen target domain. 

    Specifically, the domain-specific aggregator consists of two parts: an individual specific aggregator, and a neighbor specific aggregator. In the individual specific aggregator $\mathcal{A}_{ind} ( \cdot )$, we treat $\mathcal{M}_{ind}$ as the teacher and adapt knowledge from the source domains, enabling better generalization to the unseen domain. During training, a batch of masked domain trajectories from the source domain $X_i^{t} \in D_S^?$ is randomly sampled, and their domain-specific knowledge $\{ \mathcal{M}_{ind}^k (X_i^t) \}_{k=1}^{\vert D_S \vert}$ is queried. This knowledge is then processed by the individual specific aggregator to capture the interconnection among aggregated domain knowledge:  
    \begin{equation}
        H_i^s = \mathcal{A}_{ind} ( \sum_{k=1}^{\vert D_S \vert} \mathcal{M}_{ind}^k ( X_i^{t} ) )
    \end{equation}
    The aggregator $\mathcal{A}_{ind} ( \cdot )$ explores and combines the knowledge from multiple source domains to obtain the aggregated feature in the unseen target domain. 

    The neighbor specific aggregator $\mathcal{A}_{nei} ( \cdot )$ mirrors the design of the individual specific aggregator, with the distinction that we treat $\mathcal{M}_{nei}$ as the teacher. 
    \begin{equation}
       H_{\mathcal{E}_i}^s = \mathcal{A}_{nei} ( \sum_{k=1}^{\vert D_S \vert} \mathcal{M}_{nei}^k ( X_i^{t} ) )
    \end{equation}

    As the effectiveness of the domain-specific aggregator relies on the domain-specific extractor, it is essential to ensure sufficient training of the domain-specific extractor before proceeding with the training of the domain-specific aggregator. During the domain-specific aggregator's training, the layers associated with the domain-specific extractor should be frozen.

\subsection{Training and Inference Procedures}
\label{sec: training and inference procedure}

\subsubsection{Training Procedure}
    \begin{algorithm}
        \caption{Training Procedure}
        \label{algorithm: training procedure}
        \KwIn{source domains $D_S = \{ D_S^1, D_S^2, \cdots, D_S^K \}$ and hyperparameters $\{ \delta, e_{start}, e_{end}, \sigma, f_{low}, f_{high} \}$}

        initialize model parameters and learning rate $lr$\;
        
        // Step 1: \emph{train backbone model, domain-invariant extractor, and domain-specific extractor}\;
        \For {$e=0$ to $e_{start}$}
        {
            update parameters by optimizing Eq.(\ref{eq: step1 loss})\; 
        } 
        
        // Step 2: \emph{train domain-specific aggregator}\; 
        \For {$e=e_{start}$ to $e_{end}$}
        {
            \For {$k=1$ to $K$}
            {
                // masked domain trajectory data\;
                \If{$random() < \sigma$}
                {
                    $D_S^k \rightarrow D_S^{?}$\;
                }

                train domain-specific aggregator with learning rate $lr \times f_{high}$\;
                train other modules with learning rate $lr \times f_{low}$\; 

                update parameters by optimizing Eq.(\ref{eq: step2 loss})\; 
            }
        }

        // Step 3: \emph{train the entire method}\; 
        \For {$e=e_{end}$ to $e_{total}$}
        {
            \For {$k=1$ to $K$}
            {
                // masked domain trajectory data \;
                \If{$random() < \sigma$}
                {
                    $D_S^k \rightarrow D_S^{?}$ \;
                }
                
                train the entire method with learning rate $lr \times f_{low}$\;

                update parameters by optimizing Eq.(\ref{eq: step2 loss})\; 
            }
        }
    \end{algorithm}
    \setlength{\textfloatsep}{-1mm}
    
    By integrating the domain-invariant extractor, domain-specific extractor, and domain-specific aggregator components together, our method's training procedure is outlined in Alg.~\ref{algorithm: training procedure}. The training process consists of three steps, with a total of $e_{total}$ training epochs. 

    In the first step (Lines 3-5), we jointly train the backbone model, domain-invariant extractor, and domain-specific extractor with an initialized learning rate $lr$. The loss function, denoted as $L_{total}$, used in this step is shown as follows: 
    \begin{equation}
        L_{total} = L_{base} + \delta L_{ours}
    \label{eq: step1 loss}
    \end{equation}
    where $L_{total}$ is a combination of $L_{base}$ in the backbone model (Eq.(\ref{eq: base loss})) and our proposed loss function $L_{ours}$  (Eq. (\ref{eq: our loss})). The domain weight $\delta$ is introduced to adjust the relative weight between $L_{base}$ and $L_{ours}$, and the computation of $L_{ours}$ is shown as follows: 
    \begin{equation}
        L_{ours} = \alpha * L_{recon} + \beta * L_{diff} + \gamma * L_{similar}
    \label{eq: our loss}
    \end{equation}
    where  $L_{recon}$,  $L_{diff}$ and $L_{similar}$ represent the reconstruction loss (Eq. (\ref{eq: recon loss})), the difference loss (Eq. (\ref{eq: diff loss})), and the domain adversarial similarity loss (Eq.(\ref{eq: similar loss})), $\alpha$, $\beta$, and $\gamma$ are hyperparameters introduced to adjust relative weights between them. At the end of this step, the backbone model, domain-invariant extractor, and domain-specific extractor are well-trained.

    In the second step (Lines 7-17), we train the domain-specific aggregator until $e_{end}$ epoch.
    During this step, the learning rate of the domain-specific aggregator is set relatively high (Line 13), while the learning rate of other modules is set relatively low (Line 14). To control the learning rate, we introduce hyperparameters $f_{low}$ and $f_{high}$, representing the relatively low and high fractions compared to the initial learning rate in the first step. The loss function utilized in the second step is similar to that of the first step, differing only in the setting of a small value for $\delta^{'}$. 
    \begin{equation}
        L_{total} = L_{base} + \delta^{'} L_{ours}
    \label{eq: step2 loss}
    \end{equation}
    Additionally, we introduce a probability parameter $\sigma$, referred to as the aggregator ratio, to determine the likelihood of masking out the domain label (Lines 10-12). This is to adopt a teacher-student learning process, where the domain-specific aggregator is trained to perform similarly in the situation when the domain label is unavailable. At the end of this step, the domain-specific aggregator is well-trained.
    
    Finally, in the third step (Lines 19-28), we train the entire modules in the framework, including the backbone model, domain-invariant extractor, domain-specific extractor, and domain-specific aggregator, with a low learning rate (Line 25). Similar to the second step, we adopt a teacher-student learning process by introducing the likelihood ratio parameter $\sigma$. After this step, we obtain a well-trained multi-agent trajectory prediction model that operates effectively under the multi-source domain generalization setting.

\subsubsection{Inference Procedure}
    During the inference stage, the model receives trajectories from the unknown target domain, following the procedure depicted in Step 3 of Fig.~\ref{fig: architecture}. Each trajectory is initially processed by the selected backbone model.
    The outputs, $h_{ei}^{t, l_e}$ and $P_i$, from the backbone model, are then fed into the domain-invariant extractor, generating domain-invariant fused features $H^i$. 
    Simultaneously, these outputs also pass through individual specific extractor $\{ \mathcal{M}_{ind}^k \}_{k=1}^{\vert D_S \vert}$ and neighbor specific extractor $\{ \mathcal{M}_{nei}^k \}_{k=1}^{\vert D_S \vert}$ successively. 
    They subsequently proceed to individual specific aggregator $\mathcal{A}_{ind}$ and neighbor specific aggregator $\mathcal{A}_{nei}$ before reaching the domain specific fusion module $\mathcal{M}_{fuse}$, which produces target domain-specific fused features $H^s$. 
    Finally, the future trajectory $\hat{Y}_i^t$ is generated by combining its historical information $h_{ei}^t$ and neighbor interactions $P_i$, along with the two types of features encoding knowledge about the invariant $H^i$ and diverse patterns $H^s$ across domains. 

\section{Experiments}

    To evaluate the performance of AdapTraj, we conduct extensive experiments to answer the following research questions.
    \begin{itemize}[leftmargin=*]
        \item \textbf{RQ1}: Can our framework outperform the current state-of-the-art multi-agent trajectory prediction models under the multi-source domain generalization setting? (Sec.~\ref{subsec:exp_rq1})
        \item \textbf{RQ2}: How does our framework perform in terms of model analysis? (Sec.~\ref{subsec:exp_rq2})
        \item \textbf{RQ3}: How does each component of our method contribute to the performance? (Sec.~\ref{subsec:exp_rq3})
        \item \textbf{RQ4}: How efficient is our method in terms of inference time? (Sec.~\ref{subsec:exp_rq4})
        \item \textbf{RQ5}: How do the hyperparameters impact the performance of our framework? (Sec.~\ref{subsec:exp_rq5})
    \end{itemize}

\subsection{Experimental Setup}\label{subsec:exp_setup}
\subsubsection{Datasets}

    For this research, we carefully selected the public and the most representative datasets that are considered benchmarks in the topic of multi-agent trajectory prediction. We then utilized the datasets that contain sufficient instances and exhibit diverse trajectory characteristics/distributions to verify claim in the paper. We have endeavored to overcome limitations posed by resource availability, while showcasing applicability of our method across multiple datasets and domains. To the best of our knowledge, our research utilizes the most extensive and diverse datasets within this field of study. Finally, we conduct experiments on four real-world multi-agent interaction datasets that have been widely used in previous studies on multi-agent trajectory prediction~\cite{Evolvegraph,PECNet,TrajNet++}. Statistics of datasets are presented in Tab.~\ref{tab: details_of_datasets}, and we provide comprehensive descriptions of these datasets below: 
    \begin{itemize}[leftmargin=*]
        \item \textbf{ETH\&UCY}~\cite{ETH, UCY}: it serves as the primary benchmark for evaluating multi-agent trajectory prediction methods that contain interactions like leader-follower dynamics, collision avoidance, and group formations and dispersals. 
        \item \textbf{L-CAS}~\cite{L-CAS}: it offers valuable insights into diverse social interactions within \emph{indoor environments}, including interactions among human groups, running children, and individuals with trolleys. 
        \item \textbf{SYI}~\cite{SYI}: it is a new large-scale dataset that features significantly \emph{longer trajectories} compared to other datasets and provides long-term \emph{traffic flow changes} and \emph{complex crowd behaviors}.
        \item \textbf{Stanford Drone Dataset (SDD)}~\cite{SDD}:it is a large-scale benchmark that records trajectories in the \emph{university campus}. 
    \end{itemize}

    Apart from being extracted from various domains, the trajectories in these datasets are also recorded in different spaces and at varying time intervals. For instance, the L-CAS dataset records trajectories in world space using meters as the unit of measurement and has an interval of 0.4 seconds. In contrast, the SDD dataset captures trajectories in image pixel space with a faster interval of 1/30 seconds. To ensure a fair comparison, we follow the same data preprocessing and data splitting procedures as TrajNet++\cite{TrajNet++} for all the datasets. Specifically, we convert the trajectories to real-world coordinates and interpolate the values to obtain measurements every 0.4 seconds. Each dataset is split chronologically into train, validation, and test sets with a ratio of 6:2:2.

    For all the experiments, we select  trajectories from the SDD dataset as the target domain, while utilizing trajectories from the ETH\&UCY, L-CAS, and SYI datasets as source domains.

\subsubsection{Compared Methods}

    Recall that AdapTraj can serve as a plug-and-play module for existing models for multi-agent trajectory prediction, we employ two state-of-the-art methods as our backbone prediction models: 
    \begin{itemize}[leftmargin=*]
        \item \textbf{PECNet}~\cite{PECNet}: it incorporates the inferred distant trajectory endpoints and a non-local social layer to generate diverse yet socially compliant future trajectories. 
        \item \textbf{LBEBM}~\cite{LBEBM}: it incorporates a latent-based approach that effectively models the movement history and social context, which are crucial factors in accurately predicting multi-agent trajectories. By employing a latent space and defining a cost function, LBEBM captures the underlying patterns and relationships in the data, allowing for more accurate trajectory predictions. 
    \end{itemize}

    Given that we tackle the pioneering exploration of the multi-source domain generalization setting, we proceed with the comparison to  existing single-source domain generalization models specifically designed for multi-agent trajectory prediction. Specifically, all the training datasets are collectively treated as a single source domain to facilitate the adaptation to these models.  
    \begin{itemize}[leftmargin=*]
        \item \textbf{Counter}~\cite{Counter}: it utilizes counterfactual analysis to explore the causality between predicted trajectories and input clues, and alleviate the negative effects brought by the environment bias, i.e., remove the dependence of external factors. 
        \item \textbf{CausalMotion}~\cite{CausalMotion}: it employs an invariance loss to identify and suppress spurious correlations by capturing subtle distinctions. The method is designed to leverage only a single source domain, thus not capable of uncovering the invariant features shared by external factors that are common across multiple domains.
    \end{itemize}
    
\subsubsection{Evaluation Metrics}

    We employ two widely used evaluation metrics, namely Average Displacement Error (ADE) and Final Displacement Error (FDE), to evaluate the performance of trajectory prediction methods. Lower values of these metrics indicate superior performance: 
    \begin{itemize}[leftmargin=*]
        \item ADE measures the average Euclidean distance between the predicted locations and the corresponding ground truth locations at each time step. It quantifies the deviation between the predicted trajectory and the ground truth trajectory in terms of their overall tendencies. 
        \begin{equation}
          ADE = \sum_{t=\tau+1}^{\tau + \vert T_{pred} \vert}  \sum_{i=1}^{N(i, t)} \frac{1}{N(i, t) \times \vert T_{pred} \vert} \Vert \hat{Y}_i^t - Y_i^t \Vert ^2
        \nonumber
        \end{equation}
        \item FDE measures the Euclidean distance between the predicted final location and the ground truth final location at the final time step. It quantifies the deviation between the predicted trajectory's destination and the ground truth destination. To make the formula clear, we denote the final time step as $t_{final} = \tau + \vert T_{pred} \vert$. 
        \begin{equation}
          FDE = \sum_{i=1}^{N(i, t_{final})} \frac{1}{N(i, t_{final})} \Vert \hat{Y}^{t_{final}}_i - Y^{t_{final}}_i \Vert ^2
        \nonumber
        \end{equation}
    \end{itemize}

\subsubsection{Implementation Details}

    Following the common setting with previous studies\cite{Evolvegraph,LBEBM,PECNet,Counter}, for each given trajectory, we perform prediction for the subsequent 12 time steps (equivalent to 4.8 seconds) using observed trajectories in the previous 8 time steps (equivalent to 3.2 seconds). 
    We set the number of epochs to 300, the batch size to 32, and the value of hyperparameters $\alpha$, $\beta$, and $\gamma$ in Eq.(\ref{eq: our loss}) to 0.01, 0.075, and 0.25 for all experiments. 

\subsection{Performance Comparison (RQ1)}
\vspace{-2mm}
\label{subsec:exp_rq1}

        \begin{table*}[tbp]
        \centering
            \caption{
                Performance comparison of different methods under the multi-source domain generalization setting, with each dataset serving as the target domain and the other three datasets as the multi-source domain.
            }
            \vspace{-2.5mm}
            \resizebox{0.83\linewidth}{!}{
                \begin{tabular}{c|c|c|c|c|c|c|c|c|c|c|c}
                    \hline
                    \multirow{2}*{Backbone model} & \multirow{2}*{Learning method} & \multicolumn{2}{|c}{SDD} & \multicolumn{2}{|c|}{ETH\&UCY} & \multicolumn{2}{|c|}{L-CAS} & \multicolumn{2}{|c|}{SYI} & \multicolumn{2}{|c}{Average}\\
                     & & ADE & FDE & ADE & FDE & ADE & FDE & ADE & FDE & ADE & FDE \\
                    \hline
                    \multirow{4}*{PECNet}      & vanilla            & 0.948 & 1.785 & 0.426 & 0.617 &                                 0.282 & 0.383 & 1.113 & 1.983 & 0.692 & 1.192 \\
                                               & Counter            & 1.245 & 1.806 & 0.547 & 0.583 & 0.419 & 0.346 & 2.367 & 4.800 & 1.144 & 1.884 \\
                                               & CausalMotion       & 2.394 & 1.847 & 1.578 & 0.613 & 0.702 & 0.378 & 6.138 & 2.070 & 2.703 & 1.227 \\
                                               & AdapTraj           & \textbf{0.911} & \textbf{1.670} & \textbf{0.425} & \textbf{0.572} & \textbf{0.256} & \textbf{0.336} & \textbf{1.067} & \textbf{1.883} & \textbf{0.665} & \textbf{1.115} \\
                    \hline
                    \multirow{4}*{LBEBM}       & vanilla            & 0.829 & 1.721 & 0.340 & 0.665 &                                 0.288 & 0.519 & 1.319 & 2.663 & 0.694 & 1.392 \\
                                               & Counter            & 1.387 & 2.956 & 0.617 & 1.261 & 0.485 & 0.946 & 2.464 & 5.182 & 1.238 & 2.586 \\
                                               & CausalMotion       & 2.639 & 4.544 & 1.800 & 3.043 & 0.810 & 1.414 & 6.691 & 9.643 & 2.985 & 4.661 \\
                                               & AdapTraj           & \textbf{0.814} & \textbf{1.648} & \textbf{0.278} & \textbf{0.527} & \textbf{0.237} & \textbf{0.410} & \textbf{1.026} & \textbf{1.909} & \textbf{0.589} & \textbf{1.124} \\
                    \hline
                \end{tabular}}
                \label{tab: performance comparison RQ1-revised}
            \vspace{-6mm}
        \end{table*}

    Tab.~\ref{tab: performance comparison RQ1-revised} presents the results of all the compared methods for multi-agent trajectory prediction task under the multi-source domain generalization setting. In the experiments, Each dataset is utilized as the target domain while the other three are treated as the source domains. Specifically, we compare four baseline methods on two backbone models, namely PECNet\cite{PECNet} and LBEBM\cite{LBEBM} mentioned previously. These methods include vanilla, Counter, CausalMotion, and AdapTraj, where vanilla denotes the original implementation of the backbone model. We make several observations from the results. 

    First, LBEBM-vanilla performs better than PECNet-vanilla because it considers the movement history and social context, which are important for accurate multi-agent trajectory prediction. These findings are consistent with predictions under the setting where training and test trajectories are drawn from the same dataset. 

    Second, PECNet-Counter performs worse than PECNet-vanilla due to its counterfactual framework. This framework removes the counterfactual component capturing environmental interactions during prediction, leading to the loss of crucial information and hindering performance, especially in multi-source domain generalization. Similar patterns can be observed between LBEBM-Counter and LBEBM-vanilla.

    Third, PECNet-CausalMotion performs worse than PECNet-vanilla because the causal formalism used in CausalMotion fails to address negative transfer(demonstrated in Tab.~\ref{tab: negative transfer}), particularly in the presence of multiple source domains. The causal formalism only leverages a single source domain and cannot uncover invariant features shared across multiple domains. This issue becomes more pronounced as the number of source domains increases, resulting in poorer performance compared to directly deploying the original model on the target domain.

    Lastly, AdapTraj outperforms all baseline methods. 
    While the absolute value of improvement may be modest, such magnitude is still considered noteworthy contributions in the context of multi-agent trajectory prediction research, as demonstrated by previous studies, including but not limited to EvolveGraph~\cite{Evolvegraph}, LBEBM~\cite{LBEBM}, and PECNet~\cite{PECNet}.
    Moreover, the experiments across domains indicate that our method yields more substantial improvements in certain scenarios (e.g., SYI), while maintaining consistent superiority across all the domains. These findings further highlight the contributions of our method. 
    Building upon a new causal formulation for trajectory prediction to explicitly model four types of features: domain-invariant and domain-specific features for both the focal agent and neighboring agents. This approach is able to effectively leverage the knowledge from multiple domains and thus address the negative transfer issue for prediction in unseen target domains. The superior performance of AdapTraj is consistent across different base models, indicating its effectiveness in multi-source domain generalization for the multi-agent trajectory prediction task. 

\subsection{Model Analysis (RQ2)}
\label{subsec:exp_rq2}

    \begin{figure}
        \centerline{\includegraphics[width=.65\linewidth]{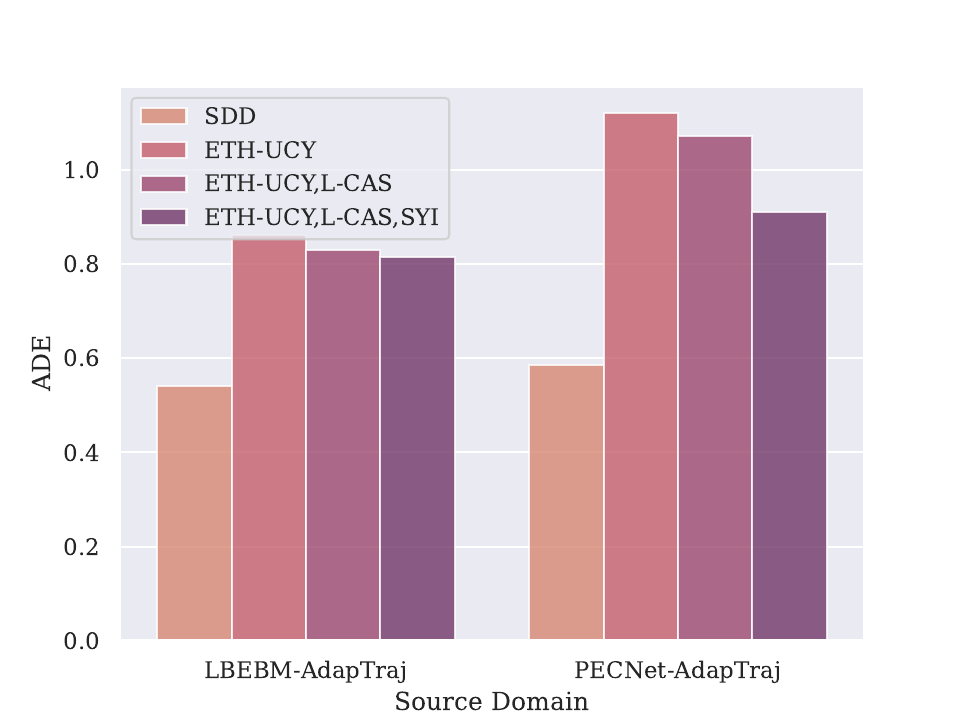}}
        \vspace{-4mm}
        \caption{Performance of AdapTraj on Various Numbers of Source Domains. }
        \label{fig: negative-ours}
    \end{figure}

    Unlike utilizing a single source domain, the main motivation for incorporating multiple source domains is the expectation that an increased number of source domains can boost performance. However, as presented by the experiments in Tab.~\ref{tab: negative transfer}, this cannot be easily achieved without specific modifications. To illustrate the effectiveness of our proposed framework,  
    we examine the performance with two backbone models on different numbers of source domains. The results are presented in Fig.~\ref{fig: negative-ours}. Overall, LBEBM-AdapTraj consistently outperforms PECNet-AdapTraj across all scenarios, which can be attributed to the enhanced ability of vanilla LBEBM in modeling the movement history and social context, which are crucial factors in accurately predicting multi-agent trajectories. Moreover, we observe that facilitated by the proposed framework, the performance improves as the number of source domains increases, which illustrates the successful mitigation of the negative transfer issue.

        \begin{table*}[tbp]
        \centering
            \caption{
                Performance comparison 
                of different methods 
                under the single-source domain generalization setting, with each dataset serving as the single-source domain and evaluated on the SDD dataset. 
            }
            \vspace{-3mm}
            \resizebox{0.75\linewidth}{!}{
                \begin{tabular}{c|c|c|c|c|c|c|c|c|c}
                    \hline
                    \multirow{2}*{Backbone model} & \multirow{2}*{Learning method} & \multicolumn{2}{|c|}{ETH\&UCY} & \multicolumn{2}{|c|}{L-CAS} & \multicolumn{2}{|c|}{SYI} & \multicolumn{2}{|c}{Average} \\
                     & & ADE & FDE & ADE & FDE & ADE & FDE & ADE & FDE \\
                    \hline
                    \multirow{4}*{PECNet}      & vanilla            & 1.203 & 1.877 & 1.901 & 2.468 &                                 1.343 & 2.093 & 1.482 & 2.146 \\
                                               & Counter            & 1.223 & 1.878 & \textbf{1.557} & 2.476 & 1.354 & 2.329 & 1.378 & 2.228 \\
                                               & CausalMotion       & 2.408 & 1.895 & 2.475 & 2.494 & 2.443 & \textbf{2.068} & 2.442 & 2.152 \\
                                               & AdapTraj           & \textbf{1.121} & \textbf{1.743} & 1.573 & \textbf{2.381} & \textbf{1.307} & 2.099 & \textbf{1.334} & \textbf{2.074} \\
                    \hline
                    \multirow{4}*{LBEBM}       & vanilla            & 0.852 & 1.798 & 1.689 & 3.200 &                                 1.087 & 2.193 & 1.209 & 2.397 \\
                                               & Counter            & 1.265 & 2.728 & 2.012 & 3.786 & 1.379 & 2.965 & 1.552 & 3.160 \\
                                               & CausalMotion       & 2.653 & 4.747 & 2.629 & 4.320 & 2.583 & 3.745 & 2.622 & 4.271 \\
                                               & AdapTraj           & \textbf{0.849} & \textbf{1.763} & \textbf{1.483} & \textbf{2.898} & \textbf{1.056} & \textbf{2.120} & \textbf{1.129} & \textbf{2.260} \\
                    \hline
                \end{tabular}}
                \label{tab: performance comparison single source-revised}
            \vspace{-5mm}
        \end{table*}

    \vspace{-1mm}
    To further compare our method and other baseline methods, 
    we vary the dataset of single-source domain and evaluate them on the SDD dataset. The results are presented in Tab.~\ref{tab: performance comparison single source-revised}. It
    can be observed that the performance of PECNet-Counter is even worse than that of the vanilla PECNet, as the counterfactual learning method disregards the influence of environments. In contrast, PECNet-AdapTraj achieves the best performance even under the single-source distribution shift setting. The experimental results indicate that our method is not only effective for multi-source scenarios, but also proves its superiority in single-source domain generalization contexts.

    \begin{table}[tbp]
    \centering
        \caption{Performance on various numbers of source domains. }
        \vspace{-3mm}
            \resizebox{0.75\linewidth}{!}{
            \begin{tabular}{c|c|c|c}
                \hline
                Method & Source Domains      & ADE & FDE \\
                \hline
                \multirow{3}*{PECNet}       & SDD                & 0.592 & 1.051 \\
                                            & ETH\&UCY            & 1.203 & 1.877 \\
                                            & ETH\&UCY, L-CAS      & 1.240 & 1.956 \\
                \hline
                \multirow{3}*{PECNet-AdapTraj}  & SDD                & 0.585 & 1.052 \\
                                            & ETH\&UCY            & 1.121 & 1.743 \\
                                            & ETH\&UCY, L-CAS      & 1.072 & 1.729 \\
                \hline
            \end{tabular}}
            \label{tab: performance comparison various source}
        \vspace{-4mm}
    \end{table}
    
    Furthermore, we demonstrate the advantages of our learning method in various settings, as depicted in Tab.~\ref{tab: performance comparison various source}. PECNet-AdapTraj slightly outperforms the vanilla PECNet even under the independent and identically distributed setting, where both training and testing are performed on the SDD dataset. However, as the distribution shift setting is considered and the number of source domains increases, the advantage of our method becomes more evident.

\subsection{Ablation Study (RQ3)}
\label{subsec:exp_rq3}

    \begin{table}[tbp]
    \centering
      \caption{ablation study with target domain SDD and source domains ETH\&UCY, L-CAS, and SYI}
      \vspace{-3mm}
            \resizebox{0.7\linewidth}{!}{
            \begin{tabular}{c|c|c|c}
                \hline
                Backbone model & Variant      & ADE & FDE \\
                \hline
                \multirow{4}*{PECNet}       & w/o specific      & 0.942 & 1.799  \\
                                            & w/o invariant       & 0.927 & 1.671 \\
                                            & ours         & 0.911 & 1.670 \\
                \hline
                \multirow{4}*{LBEBM}        & w/o specific      & 0.842 & 1.728  \\
                                            & w/o invariant       & 0.850 & 1.773 \\
                                            & ours          & 0.814 & 1.648 \\
                \hline
            \end{tabular}}
            \label{tab: ablation study}
    \end{table}

    We perform an ablation study to analyze the impact of different components in our method under the multi-source domain generalization setting. Specifically, we examined the following variants:
    \begin{itemize}[leftmargin=*]
        \item \textbf{w/o specific}: it removes domain-specific features from our method.
        \item \textbf{w/o invariant}: it removes domain-invariant features from our method.
        \item \textbf{ours}: a complete version of our framework, including both invariant and specific features.
    \end{itemize}

    The results, as presented in Tab.~\ref{tab: ablation study}, demonstrate that our framework consistently outperforms all the variants across both the PECNet and LBEBM backbone models. This indicates that the removal of any component from our framework negatively affects the overall performance. Furthermore, the contributions of these components are consistent in the two backbone models. Moreover, we make several observations from the difference between variants and vanilla models.
     
    First, the observed increase in FDE for `w/o specific' model variant compared to the vanilla model could be attributed to the difficulty of extracting effective features while neglecting domain-specific features. Previous experiments suggest that enhancing generalization performance depends on the decoupling of domain-invariant and specific features. With different backbone models, removing either module of capturing respective features in our method would potentially lead to suboptimal performance in certain metrics. Specifically, for FDE in PECNet, when domain-specific features are not explicitly captured, higher emphasis placed on learning domain-invariant features can have more negative effect on the model, thus producing even worse performance than the vanilla method.
    
    Second, the different increase in ADE for the model variants compared to the vanilla model could be attributed to the inherent model design of PECNet. PECNet does not adequately employ strict constraints on the extraction of invariant features across domains. This could lead to the unexpected incorporation of noise from domain-specific patterns into the invariant features. Compared to original PECNet, our approach further includes an explicit module that captures either domain-invariant or domain-specific features. The empirical results indicate that the module designed to extract domain-specific features (w/o invariant) is better at removing the noise of the invariant features obtained in PECNet than the module dedicated to extract domain-invariant features (w/o specific). Therefore, `w/o invariant' model variant contributes more significantly to the performance on ADE metric.

\subsection{Model Efficiency (RQ4)}
\label{subsec:exp_rq4}

    \begin{table}[bp]
    \centering
        \caption{Inference Time(Seconds) with Target Domain SDD and Source Domains ETH\&UCY, L-CAS, and SYI}
        \vspace{-3mm}
            \resizebox{0.75\linewidth}{!}{
            \begin{tabular}{c|c|c}
                \hline
                Backbone model & Learning method & Average inference time \\
                \hline
                \multirow{4}*{PECNet} & vanilla & 0.003 \\
                & Counter & 0.004 \\
                & CausalMotion & 0.003 \\
                & AdapTraj & 0.007 \\
                \hline
                \multirow{4}*{LBEBM} & vanilla & 0.027 \\
                & Counter & 0.031 \\
                & CausalMotion & 0.027 \\
                & AdapTraj & 0.030 \\
                \hline
            \end{tabular}}
            \label{tab: model efficiency}
    \end{table}

    We further evaluate the efficiency of different baselines in terms of average inference time for trajectories from the target domain. The experiments are conducted on both PECNet and LBEBM models, and the results are shown in Tab.~\ref{tab: model efficiency}. 

    The comparison of inference times among the models reveals several insights. LBEBM model exhibits a longer inference time compared to PECNet, primarily due to the higher complexity of the latent-based module. Additionally, LBEBM-Counter requires slightly more time compared to the LBEBM-vanilla because of the additional subtraction step in counterfactual prediction. On the other hand, PECNet-Counter shows a comparable inference time to the PECNet-vanilla, indicating that the inference time is primarily influenced by the backbone model. Moreover, the CausalMotion learning method demonstrates almost identical inference time to the vanilla method, as it maintains the same model architecture and inference process. Compared to the baseline learning methods, our proposed framework exhibits a slightly longer inference time. This is attributed to the utilization of both invariant and specific features, which enhances the flexibility and adaptability of the backbone model. However, it is important to emphasize that the inference time for all compared methods remains within the same order of magnitude (3-31 milliseconds). Thus, the slight differences in inference time have a negligible impact on the real-time operation of the backbone models. 
    Such a difference is unlikely to have great impact on the user experience when considering the real-time model deployment for this specific task.

\subsection{Parameter Sensitivity (RQ5)}
\label{subsec:exp_rq5}

\begin{figure}
        \centering
        \setlength{\subfigcapskip}{-2.5mm}
        \setlength{\subfigbottomskip}{0mm}
        
        \subfigure[domain weight $\delta$]{
            \includegraphics[width=.46\linewidth]{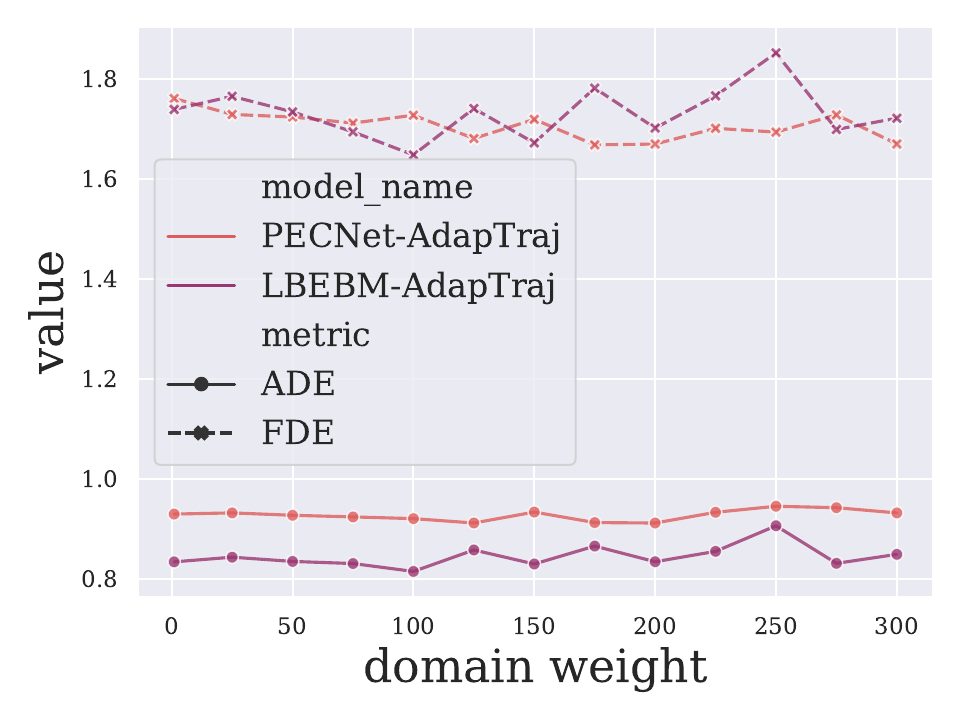}
            \label{fig: Parameter Sensitivity for domain weight}
        }
        \subfigure[aggregator start epochs $e_{start}$]{
            \includegraphics[width=.46\linewidth]{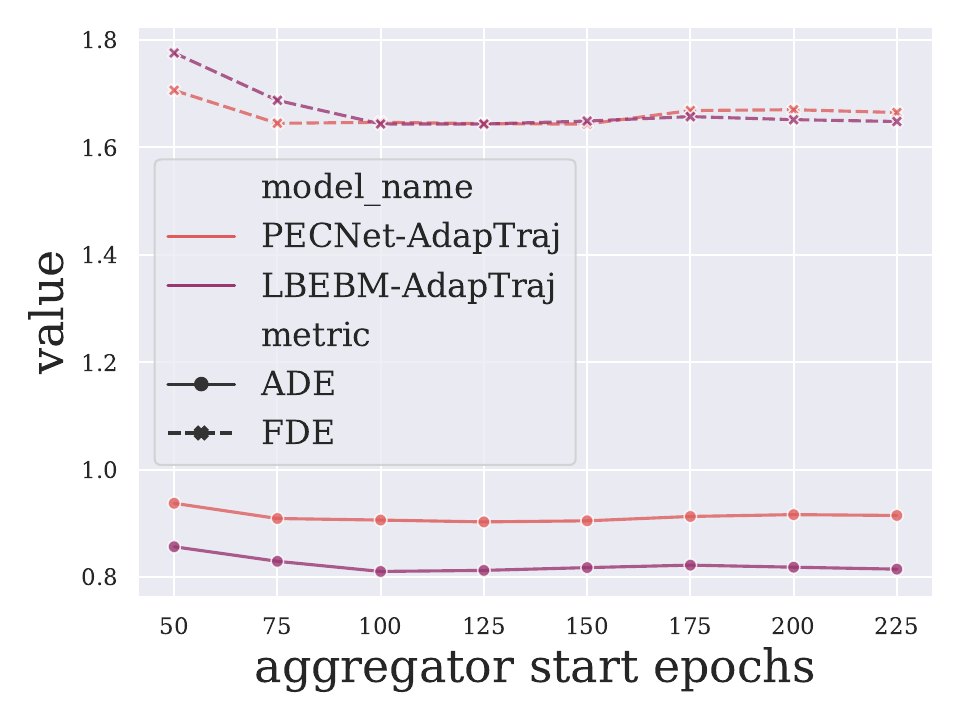}
            \label{fig: Parameter Sensitivity for aggregator_epochs}
        }
        \subfigure[aggregator end epochs $e_{end}$]{
            \includegraphics[width=.46\linewidth]{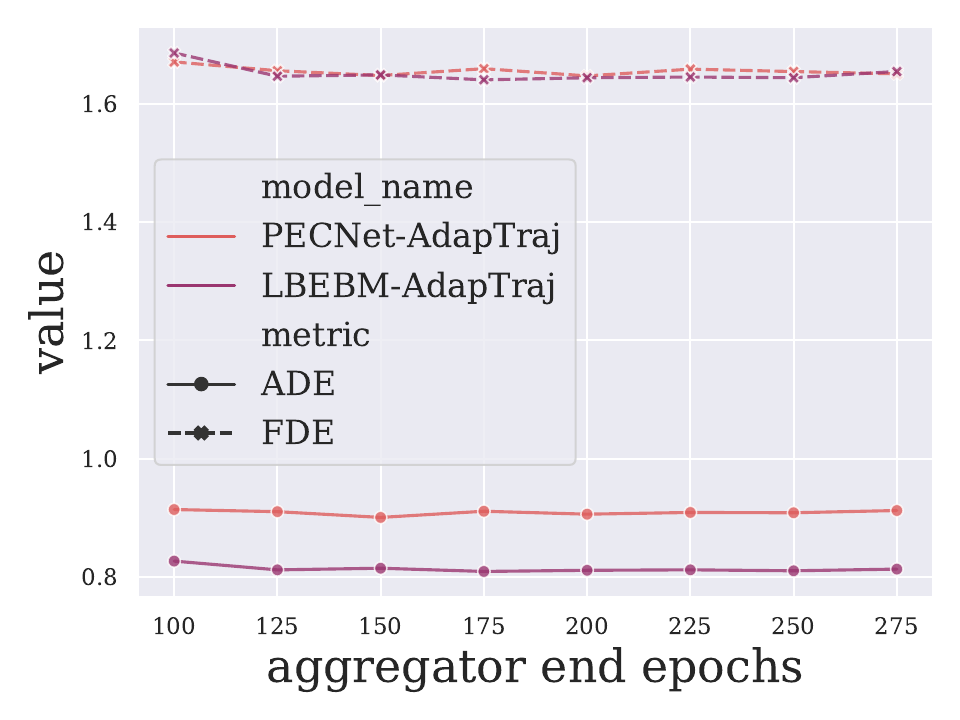}
            \label{fig: Parameter Sensitivity for aggregator_end_epochs}
        }
        \subfigure[aggregator ratio $\sigma$]{
            \includegraphics[width=.46\linewidth]{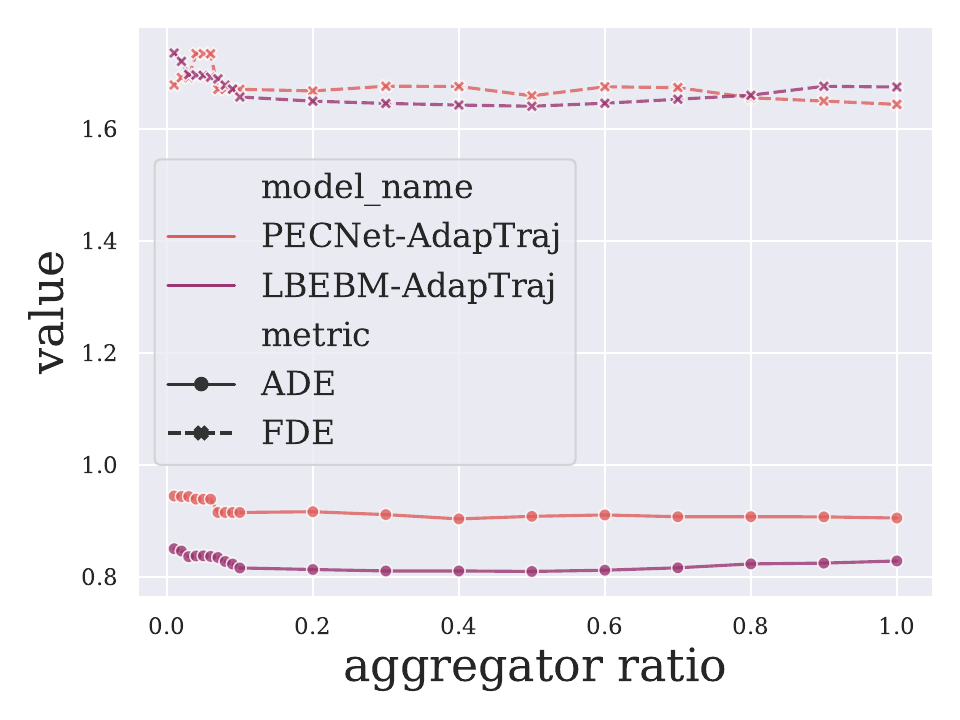}
            \label{fig: Parameter Sensitivity for aggregator_ratio}
        }
        \subfigure[low lr fraction $f_{low}$]{
            \includegraphics[width=.46\linewidth]{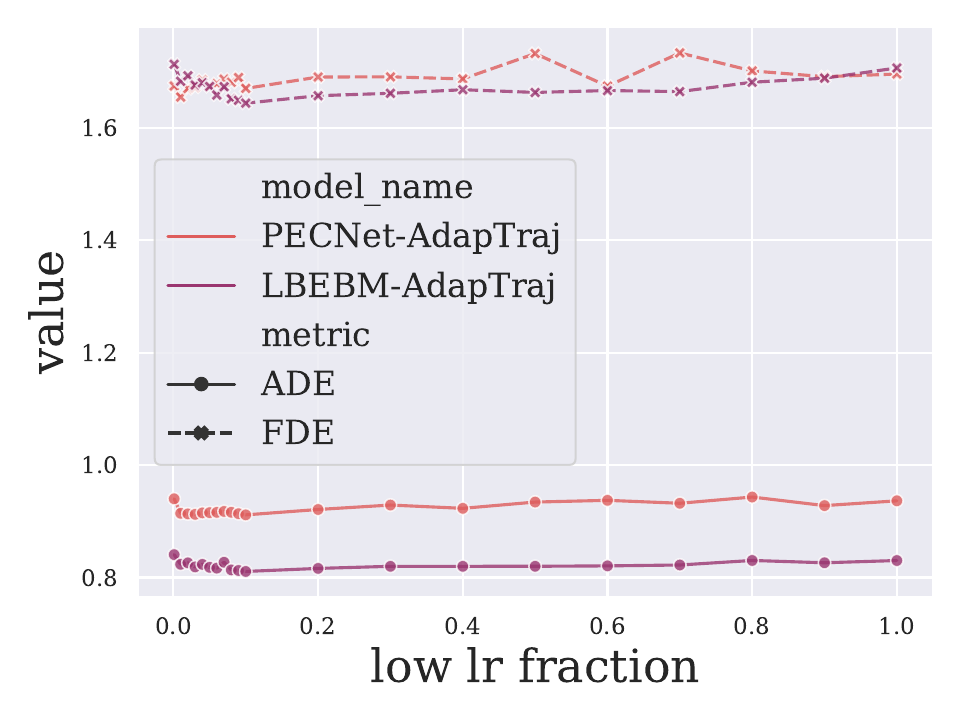}
            \label{fig: Parameter Sensitivity for low_lr_fraction}
        }
        \subfigure[high lr fraction $f_{high}$]{
            \includegraphics[width=.46\linewidth]{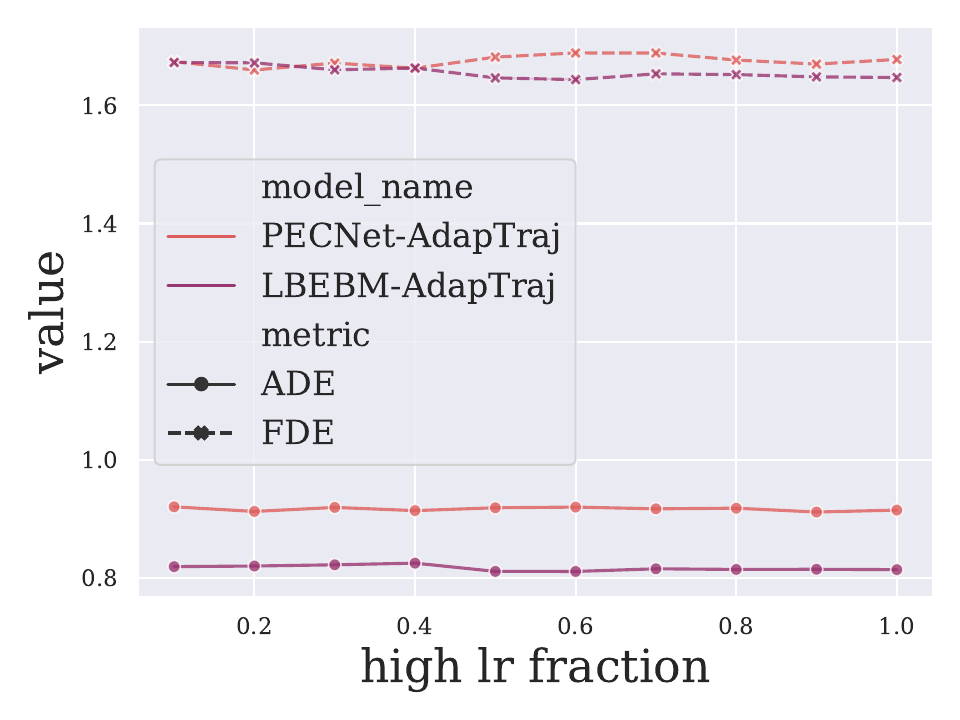}
            \label{fig: Parameter Sensitivity for high_lr_fraction}
        }
        \vspace{-1.5mm}
        \caption{
        Parameter Sensitivity Analysis. 
        }
        \label{fig: Parameter_Sensitivity}
    \end{figure}

   Considering that the number of hyperparameters is relatively large, we conduct a sensitivity analysis on the hyperparameters outlined in Alg.~\ref{algorithm: training procedure}. We employ the same experimental setting mentioned in the paper, and the experimental results are presented in Fig.~\ref{fig: Parameter_Sensitivity}. 

\subsubsection{Domain Weight $\delta$} 
    Fig.~\ref{fig: Parameter Sensitivity for domain weight} indicates that moderate values of the domain weight yield the highest performance in most cases. Extremely small domain weights hinder the mitigation of negative transfer, while excessively large domain weights impair the performance of the future trajectory generator in the backbone model.

\subsubsection{Aggregator Start Epochs $e_{start}$}
    Fig.~\ref{fig: Parameter Sensitivity for aggregator_epochs} 
    shows that a higher aggregator start epoch improves final results. Delaying the training of the domain-specific aggregator module allows the well-trained domain-specific extractor modules to enhance its effectiveness during the unseen phase. However, once the domain-specific extractor module is adequately trained, further delay in starting the training of the aggregator module does not significantly affect the final results.

\subsubsection{Aggregator End Epochs $e_{end}$}
    Fig.~\ref{fig: Parameter Sensitivity for aggregator_end_epochs} reveals a noticeable trend where a larger aggregator end epoch results in improved performance. However, once the domain-specific aggregator module has been sufficiently trained, prolonging its training does not yield significant changes in the final results.

\subsubsection{Aggregator Ratio $\sigma$}
    Fig.~\ref{fig: Parameter Sensitivity for aggregator_ratio} indicates that a larger aggregator ratio leads to better final results. 
    A higher aggregator ratio above 0.5 leads to flat or deteriorating final results, as it hinders the training of the future trajectory generator module and negatively affects performance.

\subsubsection{Low Lr Fraction $f_{low}$} 
    Fig.~\ref{fig: Parameter Sensitivity for low_lr_fraction} demonstrates a clear trend where too smaller or too larger low learning rate fraction, which hampers the regulation of the relationship between the domain-specific aggregator module and the other modules, results in poorer final results. 

\subsubsection{High Lr Fraction $f_{high}$}
    Fig.~\ref{fig: Parameter Sensitivity for high_lr_fraction} 
    shows that a larger high learning rate fraction leads to better final results, as it enables the full training of the domain-specific aggregator module. This module becomes more effective in the unseen phase, improving performance.
    
\section{Related Work}

    In this section, we first discuss existing studies on multi-agent trajectory prediction, then we present relevant work that tackles domain generalization. 
\vspace{-2mm}
\subsection{Multi-Agent Trajectory Prediction}
\vspace{-1mm}

    Multi-agent trajectory prediction problem has been widely studied in the literature, and existing studies can be roughly grouped as a combination of modeling individual mobility and interactions among agents. 

    For modeling individual mobility, various models for handling sequential data are employed as backbones to capture spatial correlations and temporal dynamics in human trajectory \cite{ICDE21-VehiclePredict,VLDB14-BehaviorPrediction}. For example, 
    PredRNN \cite{PredRNN} expands upon the inner-layer transition function of memory states in RNNs by introducing a spatial-temporal memory flow. This approach enables the joint modeling of spatial correlations and temporal dynamics at various levels within RNN. 
    Due to the more effective performance achieved by Transformer Network and Bidirectional Transformer (BERT) across different applications, they have been also adopted to replace RNN in the multi-agent trajectory prediction task~\cite{TransformerTF,ICDE21-VehiclePredict}.
   
    Interactions between agents are rather complicated since agents are governed by physical laws and social norms, such as yielding right-of-way and keeping certain social distances \cite{Evolvegraph,SocialForce,LBEBM,PECNet}. To consider these factors in human interactions, early studies \cite{SocialForce} model the interactions by hand-crafted features, such as several force terms. However, these manually designed features cannot truly reflect real-world human motion, leading to limited model performance. In this case, data-driven methods have been proposed to learn the interactions by social pooling techniques. 
    Current works use attention or graph structure to grasp complicated interactions, e.g., collective influence among agents. For instance, 
    STGAT \cite{STGAT} proposes a spatial-temporal attention network to assign different and adaptive weights to neighboring agents based on their relevance and attend to the most relevant agents to capture the collective behavior. 
    Additionally, 
    EvolveGraph \cite{Evolvegraph} proposes dynamic relational reasoning and adaptively evolving interaction graphs to account for the dynamic nature of interactions. 

    Despite prominent results achieved by these studies, they all employ classical setting of splitting data instances \emph{within one scene} for training and testing stages. 
    In contrast, our method addresses challenges of multi-agent trajectory prediction under multi-source domain generalization setting, which can serve as a solution for more use cases and scenarios in practice. 
    
\vspace{-1mm}
\subsection{Domain Genralization}
\vspace{-1mm}

    Domain generalization\cite{generalizationSurvey,generalizationSurvey2} aims to train a model that can achieve satisfactory performance on unseen target domains without accessing any prior information from them. This problem has typically been studied under two different settings, namely single-source domain generalization \cite{DSN,self-supervisedDG} and multi-source domain generalization \cite{generalizationApplication,spatialMoE}. 

    In single-source domain generalization, it is assumed that data instances from only one source domain are available. This means that the model is required to comprehend the inherent knowledge of the task and the diversity of examples from a single domain, which is promising yet quite challenging \cite{generalizationSurvey,Meta-DMoE,mixture-of-experts}. On the other hand, multi-source domain generalization reduces the difficulty in the previous setting by leveraging multiple source domains. To achieve this, numerous methods have been proposed with different design motivations, including data augmentation \cite{AdversarialDataAugmentation,ICDE22-DataAugmentation}, feature disentanglement \cite{FeatureDisentangle-SVM,FeatureDisentangle,Innovation21-AI}, domain-invariant representation learning \cite{ICDE02-ScaleInvariant}, etc. Among them, mixture-of-experts (MoE) ~\cite{mixture-of-experts} is a promising approach to improve the generalization performance. The idea of MoE is to decompose the whole training set into many subsets to be independently learned by different models. Then methods are developed to leverage the collective knowledge obtained by each expert. They either apply sparse selection methods \cite{switchTransformer,GShard,adaptiveMoE} to select a small number of experts during the inference stage or design full aggregation methods \cite{ICDE21-SoftMoE,Meta-DMoE,spatialMoE} to extract and combine the knowledge from all the experts to increase the expressive power of independent models. Drawing on the motivation of mixture-of-experts, we propose a novel aggregation module by utilizing a causal formalism in our model.

\vspace{-1mm}
\section{Conclusion}
\vspace{-1mm}

    In this work, we identify the limitations through quantitative experiments on how to obtain a generalizable model that is effective at tackling unseen examples in multi-agent trajectory prediction task. Furthermore, we propose a multi-source domain generalization framework tailored for multi-agent trajectory prediction named AdapTraj, which is adaptable to a variety of models with domain-invariant extractor, domain-specific extractor, and domain-specific aggregator to effectively leverage the knowledge from multiple domains and thus address the negative transfer issue in trajectory prediction for unseen target domains. Through extensive experiments using different domains based on four real-world datasets, we demonstrate that our proposed framework consistently outperforms other baselines by a substantial margin.

\section*{Acknowledgment}

    This work is supported by NSFC No. 62372430 and the Youth Innovation Promotion Association CAS No.2023112.
    Tangwen Qian is supported by China Scholarship Council to visit NTU, and this research was conducted during her visit.

\bibliographystyle{IEEEtran}
\bibliography{ref}

\end{document}